\documentclass[preprint,review]{elsarticle}
\makeatletter
\def\ps@pprintTitle{%
 \let\@oddhead\@empty
 \let\@evenhead\@empty
 \def\@oddfoot{\centerline{\thepage}}%
 \let\@evenfoot\@oddfoot}
\makeatother

\usepackage[utf8]{inputenc}
\usepackage[table]{xcolor}
\usepackage{pgfplotstable}
\usepackage{graphicx}
\usepackage{tikz}
\usepackage{pgfplots}
\usepackage{pgfplotstable}
\usepackage{caption}
\usepackage{subcaption}
\usepackage{amsmath}
\usepackage{amsfonts}
\usepackage{amssymb}
\usepackage{float}
\graphicspath{ {images/} }
\usepackage[english]{babel}

\begin{document}

\begin{frontmatter}

\title{WordNet2Vec: Corpora Agnostic Word Vectorization Method}

\author[wrut]{Roman Bartusiak\corref{cor}}
\ead{roman.bartusiak@pwr.edu.pl}
\author[wrut]{Łukasz Augustyniak}
\ead{lukasz.augustyniak@pwr.edu.pl}
\author[wrut]{Tomasz Kajdanowicz}
\ead{tomasz.kajdanowicz@pwr.edu.pl}
\author[wrut]{Przemysław Kazienko}
\ead{kazienko@pwr.edu.pl}
\author[wrut]{Maciej Piasecki}
\ead{maciej.piasecki@pwr.edu.pl}

\address[wrut]{Department of Computational Intelligence, Wrocław University of Science and Technology, Wrocław, Poland}

\cortext[cor]{Corresponding author}

\begin{abstract}
A complex nature of big data resources demands new methods for structuring especially for textual content. WordNet is~a~good knowledge source for comprehensive abstraction of natural language as its good implementations exist for many languages. Since WordNet embeds natural language in~the form of a~complex network, a~transformation mechanism WordNet2Vec is~proposed in~the paper. It creates vectors for each word from WordNet. These vectors encapsulate general position - role of a~given word towards all other words in~the natural language. Any list or set of such vectors contains knowledge about the context of its component within the whole language. Such word representation can be easily applied to many analytic tasks like classification or clustering. The usefulness of the WordNet2Vec method was demonstrated in~sentiment analysis, i.e. classification with transfer learning for the real Amazon opinion textual dataset.

\end{abstract}


\end{frontmatter}


\section{Introduction}

With a~fast technological growth, the ability to solve complex problems increased. More and more data continuously generated by social media and various IT systems requires more complex, accurate and efficient methods and algorithms in~order to provide valuable insight. The tools of Big Data Analytics and accessible to everyone High Performance Computing facilitate to face these challenges but simultaneously a~wide variety of new questions and problems raised and need to be addressed solely by scientists. 

There are many aspects that must be taken into account while processing large amounts of data. One of them is~a~general problem of knowledge representation for complex structures. Textual, multimedia or networked content has an unstructured nature that is~improper for application of most known analytic methods. Recently, most of the data sets come from texts, e.g. from social media, and they are directly impacted by the complex profile of natural languages. Transforming 'big data' of such kind into 'big knowledge' still remains a~great challenge. An obvious and commonly performed step in~such transformation is~'structuring'. However, how to transform a~complex nature of natural language to a~structured form?

Overall, there are two main sources of knowledge about the natural language: (1) text corpora and (2) WordNets, which are commonly verified by linguists. The first source strongly depends on the provenance and frequency of terms or n-gramms.
The latter, in~turn, reflects general snapshot of a~given natural language. The largest WordNet is~for Polish \cite{plWNBook} and English \cite{Miller:1995:WLD:219717.219748}. Simplifying, WordNet is~a~network that embeds relationships between distinct concepts-synset (synonym sets) existing the language, i.e. it captures the nature of the language. In opposite to text corpora, it also includes rarely used words and their unusual meanings, which in~corpora may either not exist or be damped by more popular words and meanings. Additionally, WordNet contains conceptual-semantic and lexical relationships linking synsets and verified by linguists. On the other hand, WordNet itself possesses a~complex network structure, which is~inappropriate for commonly used analytic and reasoning methods. 

The main problem addressed in~this paper, is~to find a~method for transformation of the complex network structure of the whole WordNet covering natural language into a~simple structured form - vectors suitable for further processing by means of known methods. In particular, the transformation should encapsulate a~position of each term in~the WordNet network towards all other words in~WordNet to preserve general WordNet knowledge about a~single word in~the whole natural language context. We propose a~new method, called \textit{WordNet2Vec}, for textual data representation in~the vector space that is~able to handle the above mentioned requirements. Based on the network of words from WordNet, we build a~word representation in~the vector space using its distance to any other word in~the network. In order to present the pair-wise word distance, the method calculates all-pairs shortest paths in~WordNet. Thanks to that any list of vectors - list of words also reflects the complex nature of the whole language encoded within these vectors. 

To demonstrate the usefulness of the proposed vectorization method, we suggested a~use case in~which any textual document is~transformed into a~list of vectors from \textit{WordNet2Vec}. Next, such representation is~applied to classification problem, namely sentiment analysis and assignment. Finally, we compare the effectiveness of the baseline method and some recently emerged approaches with high popularity such as Doc2Vec \cite{Le2014_Doc2Vec}. Since, we derive knowledge from general language database - WordNet, our method enables to build more robust knowledge representation models and to achieve very high level of efficiency, generalization and stability, especially if applied to transfer learning scenarios.

In the experimental part of our research, the proposed \textit{WordNet2Vec} vectorization method was utilized to sentiment analysis in~the Amazon product review dataset. In general, sentiment analysis of texts means assigning a~measure on how positive, neutral or negative the text is. Using \textit{WordNet2Vec} representation and a~supervised learning approach, the sentiment is~assigned to the document according to the content of document. In other words, sentiment analysis is~the process of determining the attitudes, opinions and emotions expressed within a~text. 


The rest of this paper is~organized as follows: in~Section \ref{sec:related_work} related work is~presented. Then, the new \textit{WordNet2Vec} method and other comparative methods are described in~Section \ref{sec:methods}. The experimental design and results are discussed in~Section \ref{sec:exp_results}. Finally, the presented ideas are concluded and future work directions are sketched in~Section \ref{sec:conclusions_future}. 

\section{Related Work}
\label{sec:related_work}

Representation of the knowledge is~an area of Artificial Intelligence concerned with how knowledge can be represented symbolically and manipulated in~an automated way \cite{Brachman:2004:KRR:975621}. Textual documents are unstructured intrinsically and in~order to provide their robust processing one need to involve resolving, aggregating, integrating and abstracting - via the various methodologies. From such textual data treatment we hope to obtain accurate estimation and prediction, data-mining, social network analysis, and semantic search and visualization. Derived knowledge can be represented in~various structures including:  semantic nets, frames, rules, and ontologies \cite{Russell:2003:AIM:773294}. Due to fact that the majority of machine learning and supervised learning approaches are dealing with vector space, text representation should also be placed in~a~vector space. Recent achievements in~the topics related to textual data representation will be briefly presented further part of the this section. In particular, there are recalled word embedding methods that are based on language corpuses. Then WordNet as a~source encapsulating the knowledge of natural language that is~used by the vectorization method proposed by us is~familiarized with the reader. Additionally, there are some insights in~complex network all-pair-shortest-paths (APSP) computation. Finally an introduction to the sentiment analysis is~provided, because it was used for an exemplary application of the proposed method.    

\subsection{Word Embedding Techniques}
\label{sec:word_embeddings}

\textit{Word embedding} is~the generous name for a~collection of language modelling and feature learning techniques where words from the vocabulary are mapped to vectors of real numbers. Word embedding can be equally called \textit{word vectorization method}. The mapping is~usually done to a~low-dimensional space, but depends relatively on the vocabulary size. Historically, word embedding was related to statistical processing of big corpora and deriving features from words' concurrence in~the documents.

In general word embedding techniques can be dived into two main groups. First of them is~based on probabilistic prediction approach. Such methods trains a~model, based on a~context window composed of words from a~corpus, generalizing the result into $n$ dimensional space ($n$ is~chosen arbitrary). Then a~word is~represented as a~vector in~the space and preserves a~context property, i.e. other words in~the space that are located close to each other are frequently co-occurring in~the corpus. \textit{Word2Vec} (Word-2-Vector) \cite{Mikolov_word2vec, word_2_vec_effi} is~the most successful method from this group. It is~based on skip-grams and continuous bag of words (CBOW). Given the neighboring words in~the window CBOW model is~used to predict a~particular word $w$. In contrast, given a~window size of $n$ words around a~word $w$, the skip-gram model predicts the neighboring words given the current word. 

There were other deep and recurrent neural network architectures proposed for learning word representation in~vector space before, i.e. \cite{Luong2003, Collobert2011}, but Word2Vec outperformed them and gained much more attention. 

''Count-based'' models constitiute the second group of the word embedding techniques. GloVe algorithm, presented by Pennington et al. \cite{pennington2014glove}, is~one of them. Count-based models learn their vectors based on the word co-occurrence frequency matrix. In order to shrink the size of word vectors the dimensionality reduction algorithms are applied. The main intuition for the GloVe model is~the simple observation that ratios of word-to-word co-occurrence probabilities have the potential for encoding some form of meaning. Word vectors produced by GloVe method perform very well as the solution for word similarity tasks, and is~similar to the Word2Vec approach. Lebret and Collobert \cite{Lebret2013} and Dhillon et al. \cite{Dhillon2012} proposed some other count-based models. Lebret presented a~method that simplifies the word embeddings computation through a~Hellinger PCA of the word co-occurrence matrix. Dhillon used a~new spectral method based on CCA (canonical correlation analysis)---Two Step CCA (TSCCA)---to learn an eigenword dictionary. This procedure computes two set of CCAs: the first one between the left and right contexts of the given word and the second one between the projections resulting from this CCA and the word itself. Lebret and Collobert \cite{Lebret2015} presented alternative model based on counts. They used the Hellinger distance to extract semantic representations from the word co-occurence statistics of large text corpora. 

In conclusion, Word2Vec is~a~''predictive'' model, whereas GloVe is~a~''count-based'' model \cite{baroni-dinu-kruszewski:2014:P14-1}. However, there is~no qualitative difference between predictive models and count-based models. They are different computational methods that produce a~very similar type of a~semantic models \cite{OSterlund2015, Levy2014}.

\subsection{WordNet}

WordNet is~a~large lexical database of natural language. There are separate WordNet's for many different languages, e.g. for English \cite{Miller:1995:WLD:219717.219748, wordnet}. Nouns, verbs, adjectives and adverbs are grouped into sets of cognitive synonyms (synsets), each expressing a~distinct concept. Synsets are interlinked using conceptual-semantic and lexical relations. Various types of links can be distinguished in~WordNet:
\begin{itemize}
\item There are 285,348 semantic links between synsets:
 \begin{itemize}
 \item 178,323 both hypernym and hyponym,
 \item 21,434 similarity,
 \item 18,215 both holonym and meronym,
 \item 67,376 others connections.
 \end{itemize}
\item There are 92,508 lexical links between all words:
 \begin{itemize}
 \item 74,656 derivation,
 \item 7,981 antonym,
 \item 9,871 others connections.
 \end{itemize}
 
\end{itemize}
The resulting network of meaningfully related words and concepts can be utilized in many different fields. WordNet is~also freely and publicly available for download. WordNet's structure makes it a~useful tool for computational linguistics and natural language processing.

\subsection{All-pairs Shortest Paths}


All-pairs shortest paths (APSP) is~one of the most fundamental problem in~graph theory. The objective of that task is to compute distances between all vertices in~a~graph. Computations can be done for all types of graphs, i.e. directed, undirected, weighted, unweighted, etc. The field is~extensively explored and analyzed, because of that, complexity of task is~an open problem. multiple algorithms have been proposed to solve problem. Their complexity can vary, depending on type of graph that is~taken into consideration. Current most known solution for the problem have been proposed  by Robert Floyd \cite{floyd1962algorithm}. After success in~optimization of complexity, multiple new approaches that try to minimize computational and memory complexity emerged. Simple geometrical optimization allowed to decrease complexity to $O(\frac{n^3}{log(n)})$ \cite{chan2005all}. Work of Yijie Han \cite{han2006n} exceeds, unbreakable for long time, result of $O(\frac{n^3}{log(n)})$. His solution has complexity of $O(n^3 (\frac{log(log(n))}{log(n)})^{\frac{5}{4}})$, and is~currently best one in~terms of complexity. Downside of the mentioned solution is~its theoretical complexity. Method proposed by Robert Floyd \cite{floyd1962algorithm} thanks to its simplicity can be distributed in~easiest way, thanks to what it can be easily used in~distributed environment for big datasets .

\subsection{Sentiment Analysis}

Nowadays, the most commonly used methods to sentiment analysis are the classification approaches with classifiers such as Naive Bayes \cite{pang:2002, Ye20096527, DBLP:journals/corr/abs-1305-6143}, Support Vector Machines (SVM) \cite{Maas2011, Bai2011732, Gamon:2004:SCC:1220355.1220476}, Decision Tree \cite{Schler05theimportance, Augustyniak_Entropy}, Random Forest \cite{Parmar2002} or Logistic Regression \cite{Augustyniak_Entropy}. In addition, feature selection can improve classification accuracy by reducing the high-dimensionality to a~low-dimensional of feature space. Yousefpour et al. \cite{Yousefpour2016} proposed a~hybrid method and two meta-heuristic algorithms are employed to find an optimal feature subset. 

Another family of solution to sentiment classification are neural network based approach. Socher et al. \cite{Socher_recursivedeep} proposed recursive deep model for sentiment using treebank structure of sentences. Zhang and LeCun \cite{zhang2015text} used deep learning to text understanding from character-level inputs all the way up to abstract text concepts, using temporal convolutional networks (ConvNets). What is~more, some systems leverages both hand-crafted features and word level embedding features, like Do2Vec, with the usage of classifiers such as SVM \cite{liang2015rosemerry}.

\subsection{Transfer Learning}

Transfer learning provides system ability to recognize and apply knowledge extraction (learn in~the previous tasks) to the novel tasks (in new domains) \cite{Pan:2010:STL:1850483.1850545}. Interestingly, it is~based on human behaviour during learning. We can often transfer knowledge learned in~one situation and adapt it to the new one. Yoshida et al. \cite{YoshidaHINM11} proposed a~model, where each word is~associated with three factors: domain label, domain dependence/independence and word polarity. The main part of their method is~Gibbs sampling for inferring the parameters of the model, from both labelled and unlabeled texts. Moreover, the method proposed by them may also determine whether each word’s polarity is~domain-dependent or domain-independent. Zhou et al. \cite{Zhou2015298} developed a~solution to cross-domain sentiment classification for unlabeled data. To bridge the gap between domains, they proposed an algorithm, called topical correspondence transfer (TCT). TCT is~achieved by learning the domain-specific information from different areas into unified topics. 

\section{WordNet2Vec: WordNet-based Natural Language Representation in~the Vector Space -- Word Vectors}
\label{sec:methods}

\begin{figure}[!ht]
  \centering
  \includegraphics[scale=0.15]{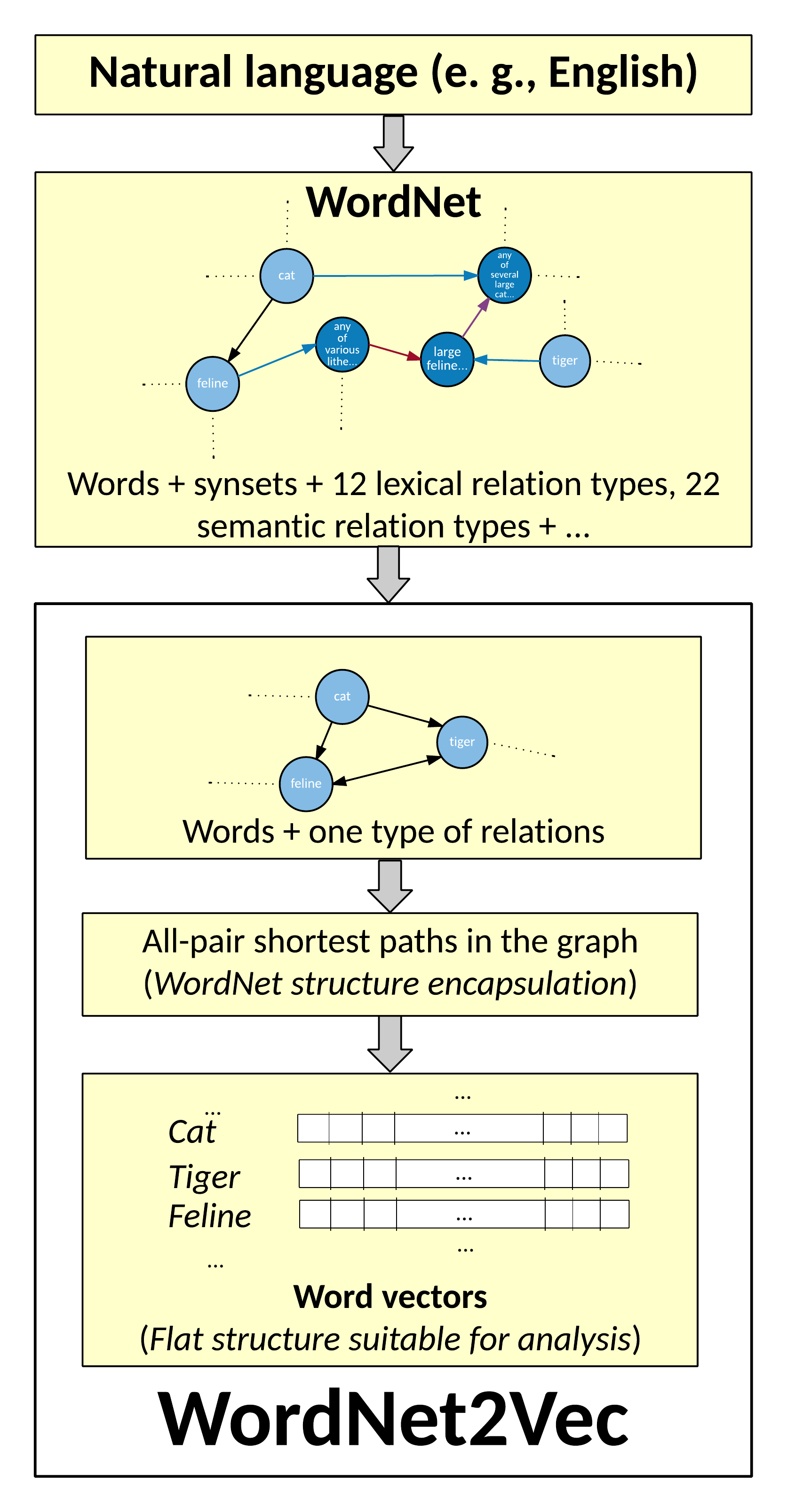}
  \caption{Architecture of WordNet2Vec method.}
  \label{fig:wordnet2vec-method}
\end{figure}

The general idea of the WordNet2Vec method is~to transfer knowledge about natural language encapsulated by the WordNet network database into word-based vector structures suitable for further processing. Its principle steps are presented in~Figure~\ref{fig:wordnet2vec-method}. 

WordNet, being developed by linguists for a~given language, consists of many synsets - meanings of words as well as lexical and semantic relations linking them. If it is~large enough -- more than a~hundred thousand synsets, it may be treated as a~reliable and comprehensive representation of the general vocabulary used by people speaking a~given language. Since WordNet has a~complex network form, it is~hardly suitable for commonly used analytic methods like reasoning by means of machine learning.

To overcome this limitation, we propose the WordNet2Vec method that provides a~set of word vectors embedding the whole WordNet. It starts with simplification of the WordNet structure into a~graph with words only and one kind of relations. The relation in~the simplified graph exists if (1) there exists a~direct lexical relation between words in~the original WordNet, (2) there exists any semantic relation between two synsets containing two considered words, (3) both words belong to one synset -- are synonyms. 

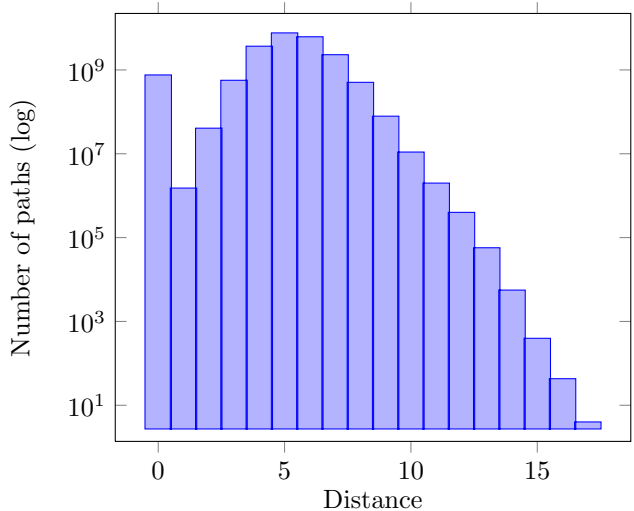
\begin{figure}[!ht]
\centering
\begin{tikzpicture}
\begin{axis}[ybar,
        ymin=0,
        enlarge y limits=0.05,
        ymode=log,
        ylabel=Number of paths (log),
        xlabel=Distance,
        ]
\addplot  table [
        col sep=comma,
        xtick=data,
        x=distance, y=count,
    ] {figures/wordnet_paths.csv};
\end{axis}
\end{tikzpicture}
\caption{WordNet shortest paths length distribution.}
\label{fig:wordnet-paths}
\end{figure}

In the next step, a~structural measure is~applied to evaluate distance from a~given node-word to any other node-word in~the simplified graph. We decided to utilize shortest paths for that purpose. It means that we had to compute all-pair shortest paths. The distribution of all pairs shortest paths is~depicted in~Figure~\ref{fig:wordnet-paths}.

Finally, for each word -- node in~the simplified network -- a~separate vector is~created. Its  coordinates correspond to shortest path lengths to all other nodes so the word vector reflects a~position of a~given word towards all other words in~the language. The set of such word vectors is~the output of the method and can be used for further processing. Such word vector set encompasses the knowledge about a~given natural language, in~particular about relations between words.

\section{WordNet2Vec Implementation -- Distributed Calculation of All-pairs Shortest Paths in~the Simplified WordNet Graph}

To demonstrate the WordNet2Vec method, it was applied to English WordNet \cite{Miller:1995:WLD:219717.219748}. We have created a~simplified network based on semantic and lexical relations present in~WordNet. Hence, we received a~graph composed of 147,478  words interconnected by 1,695,623 links.

The simplified network was utilized to compute all-pair shortest paths in~this network, so we obtained $147,478^2$, i.e. over 21 billion path lengths. 

Because of the size of the simplified network as well as computational and memory complexity of the path calculation task, we have decided to use heterogeneous computational cluster as a~environment of our experiments. Nevertheless, further optimization had to be done. In our approach, we have used distributed implementation of Dijkistra \cite{dijkstra1959note} algorithm available in~our SparklingGraph library \cite{sparkling-graph}. Most the known solutions for the all-pair shortest paths (APSP) problem has memory complexity of $O(n^3)$, where $n$ is~the number of nodes in~the graph \cite{floyd1962algorithm}. Because of that, we had to do the computations in~the iterative way. We have used 'divide and conquer approach' in~order to split APSP into smaller problems that can be computed efficiently on our cluster. In each iteration, we computed shortest paths for $1000$ vertices to all vertices in~the simplified graph. Afterwards, the results were gathered into a~coherent set of word vectors that represented distances between words in~terms of graph topology. 





\section{Use Case: WordNet2Vec Application to Sentiment Analysis}
\label{sec:usecase}

Among a~wide variety of possible application areas, the usage of WordNet2Vec method will be presented with sentiment analysis use case. In general, sentiment analysis of the texts consist in~assigning a~positive, neutral or negative measure the text. With usage of \textit{WordNet2Vec} representation we use a~supervised learning approach to generalize the sentiment classes assigned to the document. Then, according to the content of new document, it is~possible with a~trained model to infer its sentiment class. In such a~scenario one can determine the attitudes, opinions and emotions expressed within a~text. 

\begin{figure}[!ht]
  \centering
  \includegraphics[scale=0.13]{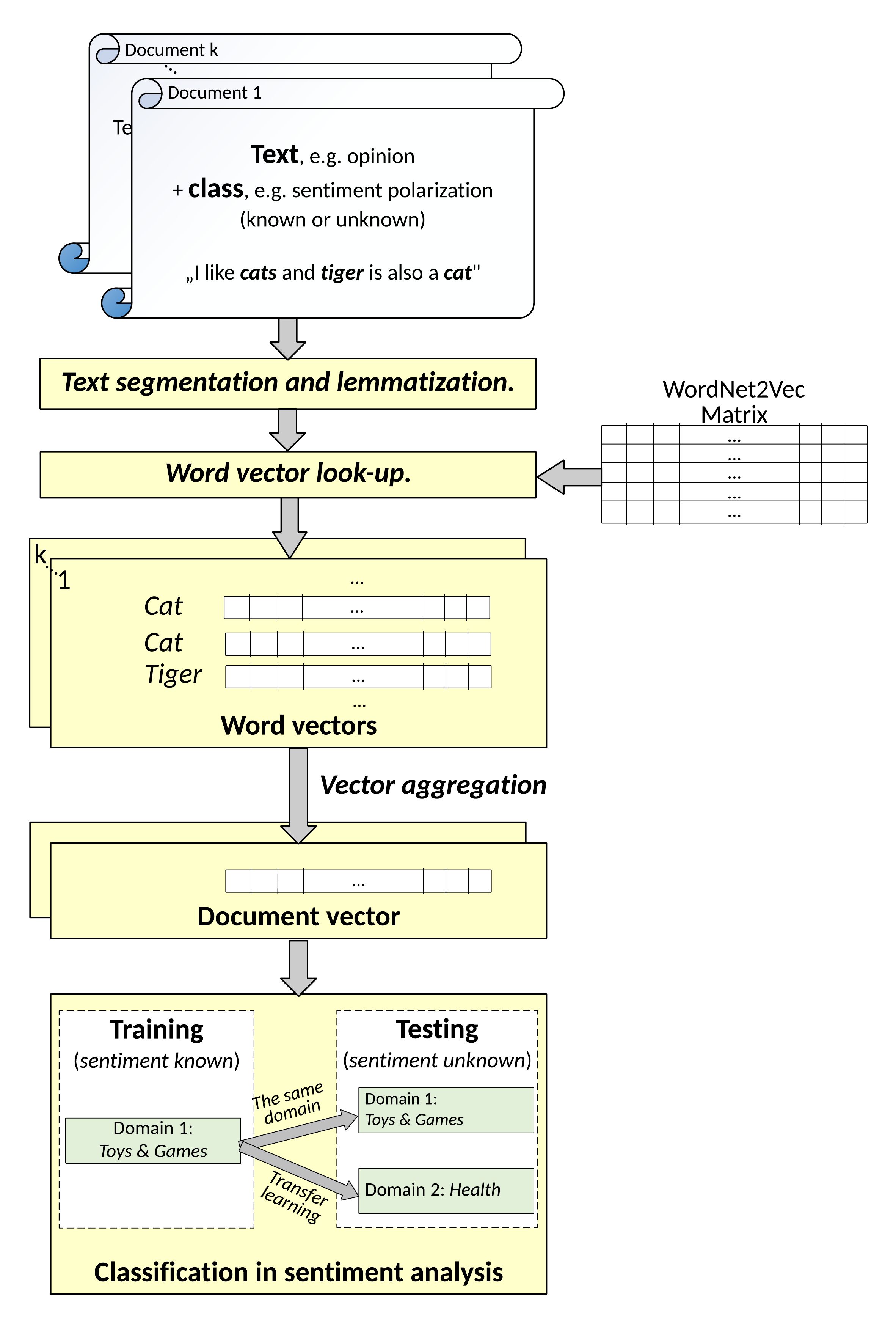}
  \caption{Exemplary application of WordNet2Vec method to Sentiment Analysis problem.}
  \label{fig:usecase}
\end{figure}

In details, the sentiment analysis can be performed with appropriate sequence of processing steps that includes: text segmentation and lematization, vector look-up in~the WordNet2Vec matrix for each word in~the document, aggregation of vectors for all words within documents, train/test dataset split and finally classifier learning and testing (ex. within the same domain or across domains - transfer learning). In order to accomplish the learning and testing phase properly we should have some sentiment classes assigned to all documents. The overall flow of the sentiment assignment is~presented in~Figure~\ref{fig:usecase} and all of the steps are discussed bellow.  

\subsection{Text segmentation and lematization}

Firstly, some basic natural language processing methods must be used: segmentation and lemmatization. In order to process each document has to be segmented into words. Due to the fact that WordNet2Vec matrix is~prepared for words in~their lemma form, each word from each document must be lemmatized. It has to be emphasized here that the method has a~disadvantage as it works only for words that are present in~WordNet. However, due to the fact that WordNet is~authoritative and reliable representation of language, it may be suitable to almost all possible application domains.

\subsection{Vector look-up in~the WordNet2Vec Matrix for a~word}

The method for word vectorization proposed in~the paper provides pre-computed WordNet2Vec Matrix. It contains vector representation for each word from WordNet. In the next step of the flow, vectors for all words for all documents are retrieved from mentioned matrix with  $O(1)$ time.

\subsection{Aggregation of vectors for all words within documents}

Due to the fact that sentiment assignment is~performed for documents, they have to be represented in~single vector. Some aggregation over all words from the document must be applied. In order to compute single vector that will represent document, we are proposing to sum up vectors of each word that appears in~the document, see Eq. \ref{eq:document-vectorization}),
\begin{equation}\label{eq:document-vectorization}
v(d)=\sum_i^{||d||}\overrightarrow{v(d_i)}
\end{equation}
where $\overrightarrow{v(\cdot)}$ represents a~vector from WordNet2Vec Matrix, $||d||$ denotes a~number of lemmas in~a~document and $d_i$ is~an $i$th lemma from document $d$.

\subsection{Dataset split, classifier learning and testing}

Once, all the documents possess a~single vector representation, they form a~dataset appropriate for classifier training and testing. In order to examine classifiers generalization abilities it is~proposed to accomplish two distinct scenarios: learning and testing within the same domain of the text (same type of documents) or transfer learning and testing (learning on one domain and testing on totally another one).  

\section{Experiments and Results}
\label{sec:exp_results}

Following the scenario presented in~the Section \ref{sec:usecase}, we performed validity tests by examining sentiment assignment task. In the following sections we describe the data that utilized, the details of experiments and received results.

\subsection{Datasets}
\label{sec:dataset}

We chose data from one data source - Amazon e-commerce platform \cite{McAuley:2013:HFH:2507157.2507163}. The data spans over a~period of 18 years, including ~35 million reviews up to March 2013. Reviews include product and user information, ratings, and a~plain text review. Some basic statistical information about the dataset are presented in~Table~\ref{tab:amazon-movies}.

\begin{table}[t]
\renewcommand{\arraystretch}{1.3}
\caption{Statistics of the whole Amazon Reviews Dataset.}
\label{tab:amazon-movies}
\centering
\begin{tabular}{c|c}
\hline
Number of reviews & 34,686,770 \\ \hline
Number of users   & 6,643,669 \\ \hline
Number of products & 2,441,053 \\ \hline
Users with $>$ 50 reviews & 56,772 \\ \hline
Median no. of words per review & 82 \\\hline
Time span & Jun 1995 - Mar 2013
 \\\hline
\end{tabular}
\end{table}

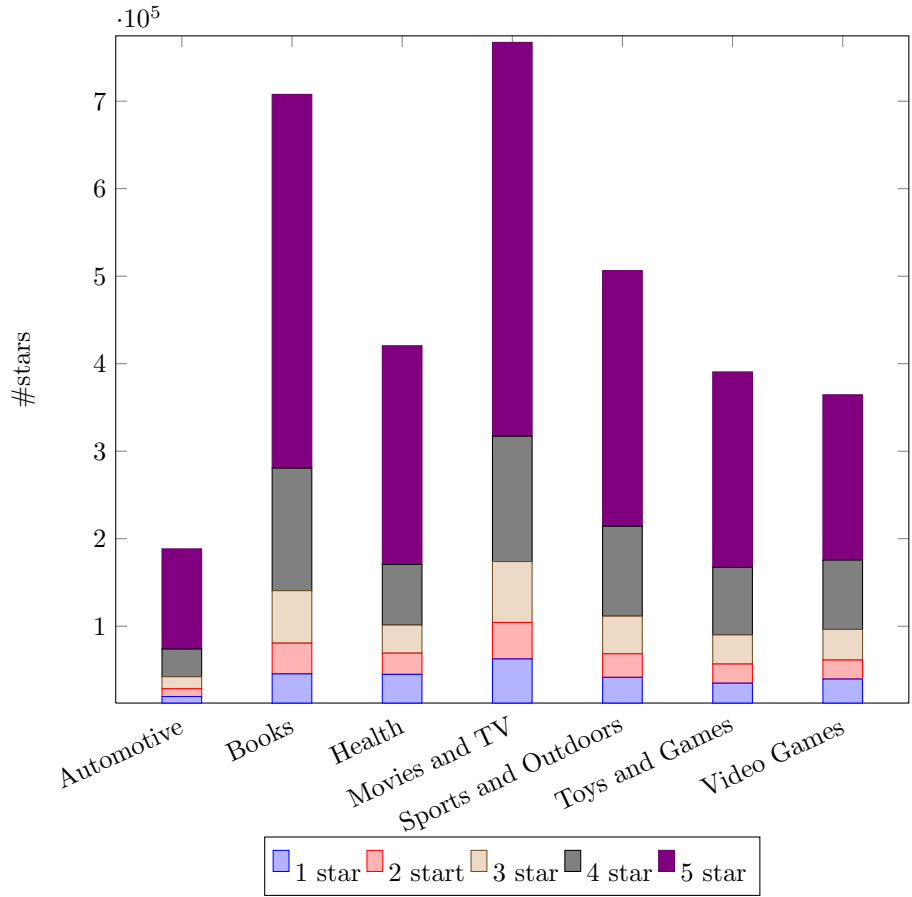
\begin{figure}[!ht]
  \centering

\begin{tikzpicture}
\begin{axis}[
    ybar stacked,
	bar width=15pt,
   enlarge y limits=0.01,
   width=\textwidth,
    legend style={at={(0.5,-0.20)},
      anchor=north,legend columns=-1},
    ylabel={\#stars},
    symbolic x coords={Automotive,Books,Health,Movies and TV,Sports and Outdoors, Toys and Games, Video Games},
    xtick=data,
    x tick label style={rotate=25,anchor=east,yshift=-7pt},
    ]
\addplot+[ybar] plot coordinates {(Automotive,19688) (Books,45532)  (Health,44972) (Movies and TV,62619) (Sports and Outdoors,41599) (Toys and Games,34911) (Video Games,39619)};
\addplot+[ybar] plot coordinates {(Automotive,8944) (Books,35160)  (Health,24272) (Movies and TV,41701) (Sports and Outdoors,26890) (Toys and Games,21884) (Video Games,21752)};
\addplot+[ybar] plot coordinates {(Automotive,13645) (Books,59946)  (Health,32088) (Movies and TV,69565) (Sports and Outdoors,43009) (Toys and Games,33287) (Video Games,34947)};
\addplot+[ybar] plot coordinates {(Automotive,31621) (Books,140017)  (Health,69229) (Movies and TV,143248) (Sports and Outdoors,102785) (Toys and Games,77115) (Video Games,79270)};
\addplot+[ybar] plot coordinates {(Automotive,114486) (Books,427093)  (Health,249946) (Movies and TV,450031) (Sports and Outdoors,292051) (Toys and Games,223395) (Video Games,188878)};
\legend{\strut 1 star, \strut 2 start, \strut 3 star, \strut 4 star, \strut 5 star}
\end{axis}
\end{tikzpicture}

  \caption{Distribution of original scores expressed in~stars over domains in~Amazon Dataset.}
  \label{fig:hist_stars}
\end{figure}

The whole experiment was conducted on selected part of the Amazon data that consisted of 7 domains, namely: Automotive, Sports and Outdoors, Books, Health, Video Games, Toys and Games, Movies and TV. Due to the fact that the distribution of classes is~important while interpreting the results of classification validation, the proper histogram is~presented in~Figure~\ref{fig:hist_stars}. The domains of review dataset that were chosen for the experiment are listed in~Table~\ref{tab:domains}).

\begin{table}[H]
\centering
\caption{Dataset's domains used in~experiment.}
\label{tab:domains}
\begin{tabular}{c|c}
\hline
Domain & Number of reviews \\
\hline
Automotive & 174,414 \\
Book & 697,225 \\
Health & 428,781 \\
Movie TV & 765,961 \\
Sports Outdoor  & 504,773 \\
Toy Game & 389,221 \\
Video Game & 364,206 \\
\hline
\end{tabular}
\end{table}

In order to check the accuracy of the proposed methods, we extracted the sentiment orientation from ratings expressed with stars. Ratings were mapped to the following classes: ''positive'', ''neutral'' and ''negative'', using 1~and 2~stars, 3~stars, 4~and 5~stars respectively, see Table~\ref{tab:stars}.

\begin{table}[H]
\renewcommand{\arraystretch}{1}
\caption{Star rating mapping to sentiment classes.}
\label{tab:stars}
\centering
\begin{tabular}{c|c}
\hline
\textbf{Star Score} & \textbf{Sentiment Class}\\
\hline
\includegraphics[width=10px]{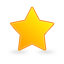} & Negative \\  
\includegraphics[width=10px]{figures/star.png}\includegraphics[width=10px]{figures/star.png}   & Negative \\
\hline
\includegraphics[width=10px]{figures/star.png}\includegraphics[width=10px]{figures/star.png}\includegraphics[width=10px]{figures/star.png} & Neutral \\  
\hline
\includegraphics[width=10px]{figures/star.png}\includegraphics[width=10px]{figures/star.png}\includegraphics[width=10px]{figures/star.png}\includegraphics[width=10px]{figures/star.png} & Positive\\  
\includegraphics[width=10px]{figures/star.png}\includegraphics[width=10px]{figures/star.png}\includegraphics[width=10px]{figures/star.png}\includegraphics[width=10px]{figures/star.png}\includegraphics[width=10px]{figures/star.png} & Positive\\
\hline
\end{tabular}
\end{table}


\subsection{Experimental setup}

Our experiments were divided into two distinct groups. First of them consisted of classical machine learning evaluation using train/test split done on each of seven mentioned datasets. In the second group of experiments we used one vs all transfer learning evaluation. Whenever one domain was used as a~training set, it was evaluated on the rest of domains. Thanks to that we experienced the quality of both, classical sentiment analysis task based on given domain dataset, and on transfer learning between different domains. In both experiment groups we have used two methods of text vectorization: proposed by us WordNet2Vec and as a~reference - Doc2Vec. 

Doc2Vec is~the generalization of Word2Vec algorithm to document level. Word2Vec model shows how a~word usually is~used in~a~window context according to other words (how words co-occur with each other). The procedure of counting Doc2Vec is~very similar to Word2Vec, except it generalizes the model by adding a~document vector. There are two methods used in~Doc2Vec: Distributed Memory (DM) and Distributed Bag of Words (DBOW). The first one attempts to predict a~word given its previous words and a~paragraph vector. Even though the context window moves across the text, the paragraph vector does not (hence distributed memory) and allows for some word-order to be captured. On the other hand, DBOW predicts a~random group of words in~a~paragraph given only its paragraph vector. In our experiments, we used Distributed Memory method trained on the Amazon SNAP (see Section \ref{sec:dataset}) review dataset. The separate model was trained for each domain and length of a~vector equal to 400. 

In order to evaluate standard classification approach we additionally provide a~baseline~$F1_{weighted}$ value that would be achieved if we would use a~classifier that returns always a~class that is~a~major one. 

Logistic regression was selected as a supervised learning model and in order to train it we have used limited memory BFGS algorithm \cite{liu1989limited}. It was a~trade-off for computation vectorized documents in the space size dependent on Wordnet2Vec Matrix size. Due to the fact that datasets used in the experiments are imbalanced, see Figure~\ref{fig:hist_stars}), we express the evaluation result using appropriate measure - weighted F1 score (Equation \ref{eq:f1-weighted}). In order to compare two approaches (WordNet2Vec and Doc2Vec), we have used statistical test on paired measures for each of experiments. We have used Wilcoxon signed-rank test \cite{randles1988wilcoxon} with confidence level $\alpha=0.05$ (Tables \ref{tab:wilcoxon-classification} and \ref{tab:wilcoxon-transfer}). To provide deeper insight of differences between results achieved by methods, we are presenting also histograms of differences between $F1_{weighted}$ of both methods (see Figures \ref{fig:classification-diff}, \ref{fig:transfer-diff}).

 \begin{eqnarray}
  \label{eq:f1-weighted}
 F1_{weighted} & = & \frac{ \sum_i^k||c_i||*F1_{c_i}}{\sum_i^k||c_i||} \\
 k & - & \text{number of classes} \nonumber  \\
 c_i & - & \text{classification results for class $i$} \nonumber  \\
 F1_{c_i} & - & \text{$F1$ score for $c_i$ classification results} \nonumber
 \end{eqnarray}
 
  \begin{equation}
 F1  =  2 * \frac{precision*recall}{precision+recall}
 \label{eq:f1}
 \end{equation}

\newcommand{\classifierResults}[3] {
\begin{tikzpicture}
\begin{axis}[ybar,
width=\textwidth,
	 symbolic x coords={Doc2Vec,WordNet2Vec},
	ylabel=$F1_{weighted}$,ymin=0,ymax=1,
enlarge x limits=0.9,
height=100,
x tick label style={rotate=21, anchor=east, align=center},
	grid=major,
	xtick={Doc2Vec,WordNet2Vec}]
\addplot 
	coordinates {(Doc2Vec,#1) }; 
\addplot 
    coordinates {(WordNet2Vec,#2) };	 
    
 \draw[very thick,red,dashed] (rel axis cs:0,#3) -- node[above=2pt,fill=none] {Baseline} (rel axis cs:1,#3);
\end{axis}

\end{tikzpicture}
}

\subsection{Generalization ability with regards to in-domain classification}

\begin{figure}[H]
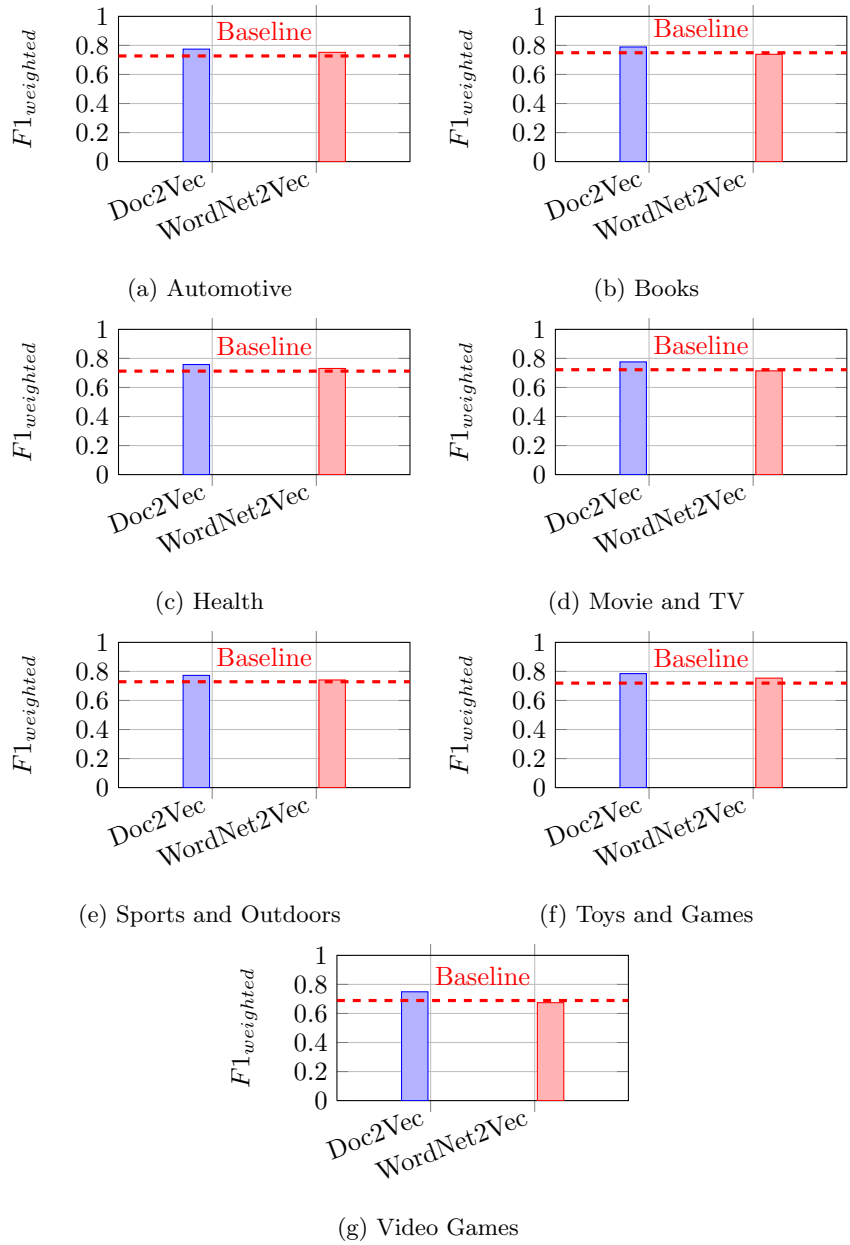

    \centering
    \vspace{-1.07\baselineskip}
    \begin{subfigure}[b]{0.45\textwidth}
        \classifierResults{0.7743242776707097}{0.75213430921958}{0.72744247808769802}
        \vspace{-1.07\baselineskip}
        \caption{Automotive}
        \label{fig:classifying-Automotive}
    \end{subfigure}
    ~ 
    \begin{subfigure}[b]{0.45\textwidth}
        \classifierResults{0.7891206724415528}{0.7387924459689805}{0.74945726253332901}
        \vspace{-1.07\baselineskip}
        \caption{Books}
        \label{fig:classifying-Book}
    \end{subfigure}

    \begin{subfigure}[b]{0.45\textwidth}
        \classifierResults{0.7579629778865241}{0.7306991649175504}{0.71242609451237404}
        \vspace{-1.07\baselineskip}
        \caption{Health}
        \label{fig:classifying-Health}
    \end{subfigure}
    ~ 
    \begin{subfigure}[b]{0.45\textwidth}
        \classifierResults{0.7758454700115609}{0.7142791959539099}{0.72212881229937631}
        \vspace{-1.07\baselineskip}
        \caption{Movie and TV}
        \label{fig:classifying-Movies}
    \end{subfigure}
    
    \begin{subfigure}[b]{0.45\textwidth}
        \classifierResults{0.7724441494169736}{ 0.7410638537857464}{0.72884645105074064}
        \vspace{-1.07\baselineskip}
        \caption{Sports and Outdoors}
        \label{fig:classifying-Sport}
    \end{subfigure}
    ~
        \begin{subfigure}[b]{0.45\textwidth}
        \classifierResults{0.7848331552899172}{0.7537304795247867}{0.71938996161546021}
        \vspace{-1.07\baselineskip}
        \caption{Toys and Games}
        \label{fig:classifying-Toy}
    \end{subfigure}
    
            \begin{subfigure}[b]{0.45\textwidth}
        \classifierResults{0.7491520802152264}{0.6740924153656368}{0.68902513218533878}
        \vspace{-1.07\baselineskip}
        \caption{Video Games}
        \label{fig:classifying-Video}
    \end{subfigure}

    \caption{Train/test classification results.}\label{fig:classifying}
\end{figure}

\begin{figure}[H]
\centering
\begin{tikzpicture}
\begin{axis}[
    ybar,
    ymin=0,
    ylabel=Count,
    xlabel=Difference,
    tick label style ={/pgf/number format/fixed},
]
\addplot +[
    hist={
        bins=7,
        data min=-0.1,
        data max=0
    }   
] table [y index=0] {classification_wordnet_vs_doc2vec.csv};
\end{axis}
\end{tikzpicture}
\caption{Histogram of difference between $F1_{weighted}$ measure for WordNet2Vec and Doc2Vec based classifiers in~train/test experiments.}
\label{fig:classification-diff}
\end{figure}
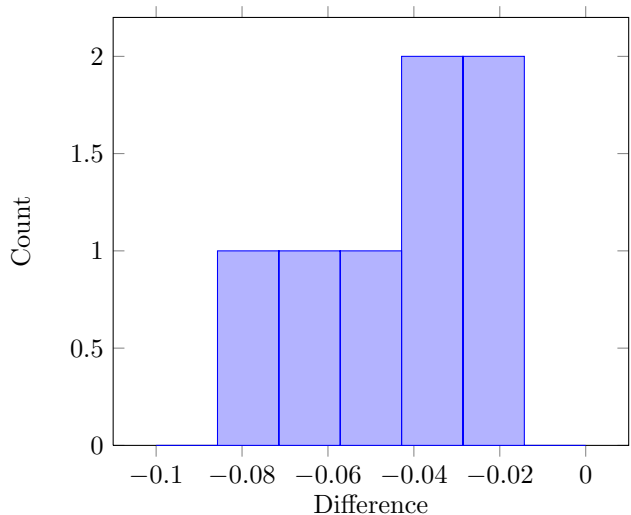

For every domain from group of seven all together, that was examined in the experiments, Doc2Vec is~slightly better then proposed the approach (Figure~\ref{fig:classifying}). The difference between the result achieved by Doc2Vec and WordNet2Vec is~not so huge what can be observed in Figure~\ref{fig:classification-diff}. Nevertheless, it is~statistically significant, what was shown by Wilcoxon rank-sum test (Table~\ref{tab:wilcoxon-classification}). It is~important to notice that statistical analysis of results showed also that WordNet2Vec is~not worse than baseline. 

\begin{table}[H]
\renewcommand{\arraystretch}{1.3}
\caption{Wilcoxon rank-sum test results for classification.}
\label{tab:wilcoxon-classification}
\centering
\begin{tabular}{l|c|c|r}
Method 1 & Method 2 & $H_a$ & p-value \\ \hline
Doc2Vec & WordNet2Vec & $F1_{Doc2Vec}>F1_{WordNet2Vec}$ & 0.007813 \\
WordNet2Vec & Baseline & $F1_{WordNet2Vec} \neq F1_{Baseline}$  & 0.2969 
\end{tabular}
\end{table}

\subsection{Generalization ability with regards to transfer learning}

\begin{figure}[H]
\centering
\includegraphics[width=0.6\textwidth]{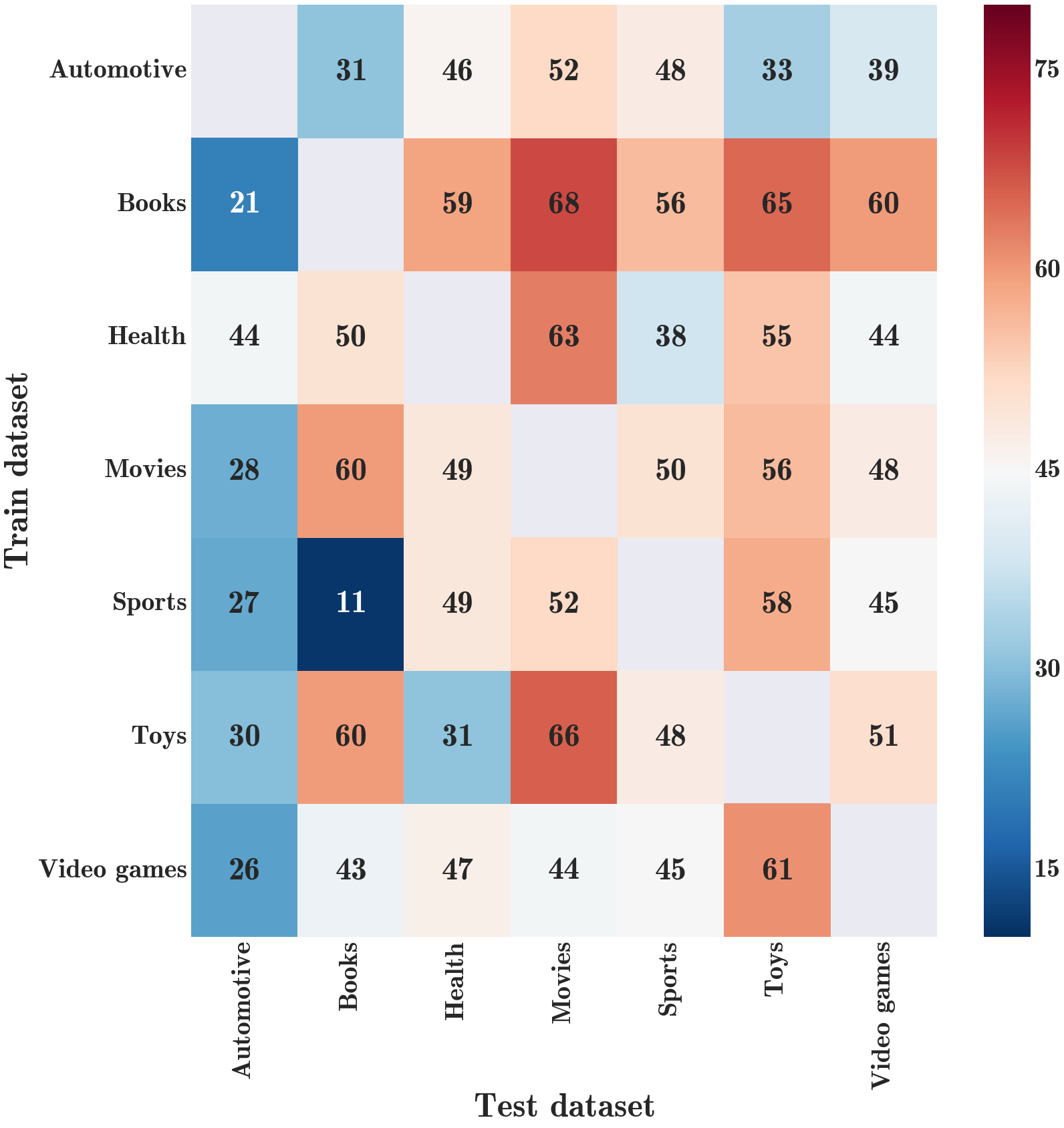}
           \caption{$F1_{weighted}$ expressed in $\%$ in~transfer learning using Doc2Vec.}
        \label{fig:transfer-doc}
    \end{figure}

\begin{figure}[H]
\centering
\includegraphics[width=0.6\textwidth]{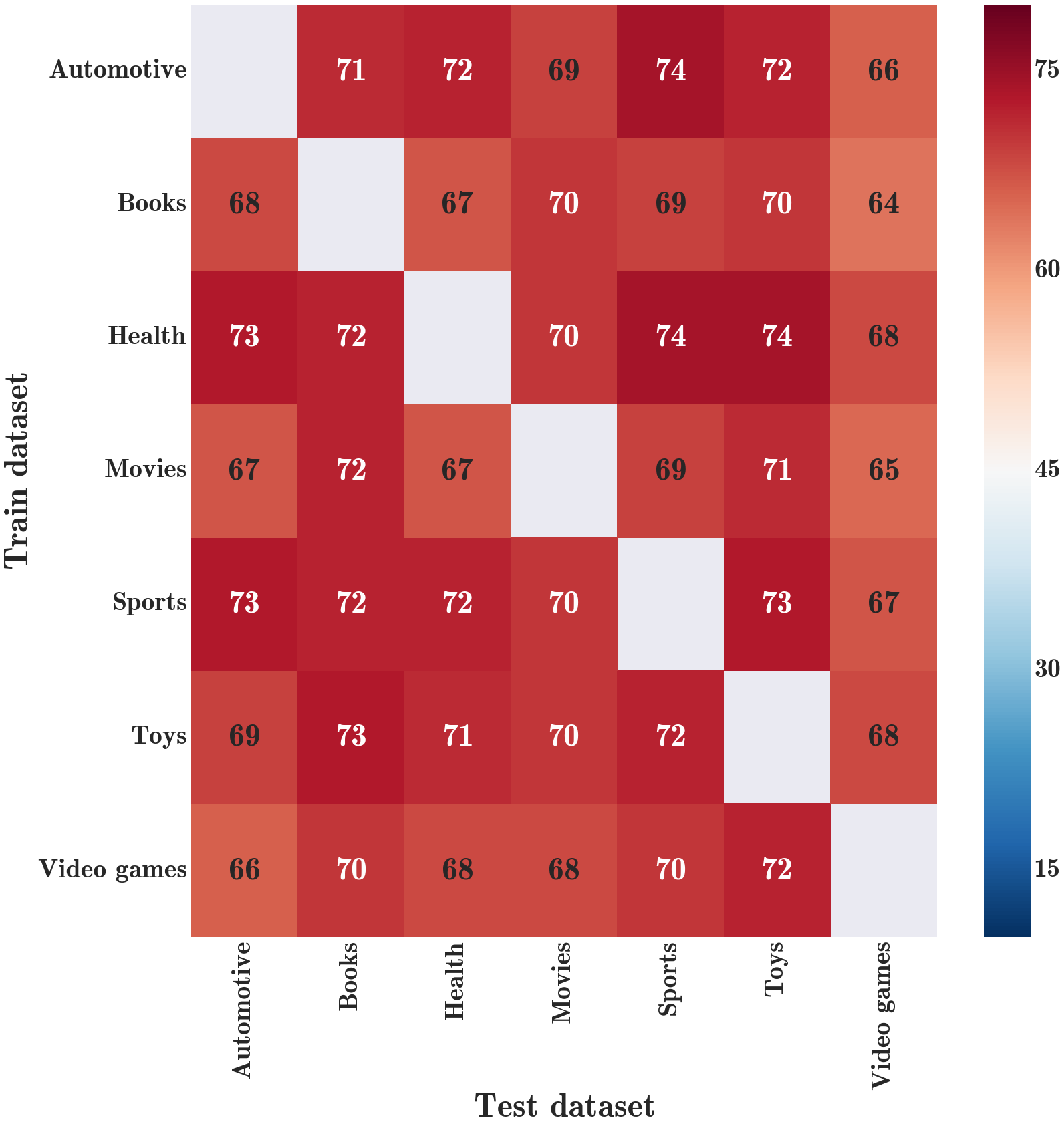}
        \caption{$F1_{weighted}$ expressed in $\%$  in~transfer learning using WordNet2Vec.}
        \label{fig:transfer-wordnet}

    \end{figure}

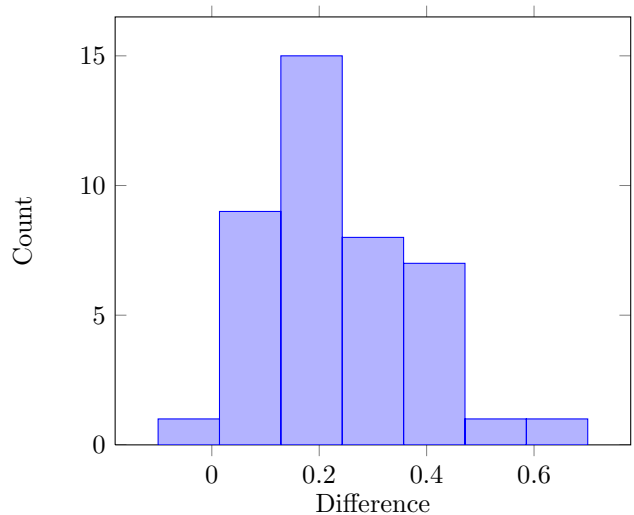
\begin{figure}[H]
\centering
\begin{tikzpicture}
\begin{axis}[
    ybar,
    ymin=0,
    ylabel=Count,
    xlabel=Difference,
]
\addplot +[
    hist={
        bins=7,
        data min=-0.1,
        data max=0.7
    }   
] table [y index=0] {transfer_wordnet_vs_doc2vec.csv};
\end{axis}
\end{tikzpicture}
\caption{Histogram of differences between F1 Weighted measure for WordNet2Vec and Doc2Vec based classification in~transfer learning experiments.}
\label{fig:transfer-diff}
\end{figure}

\begin{figure}[H]
\centering
\begin{tikzpicture}
\begin{axis}[
    ybar,
    ymin=0,
    ylabel=Count,
    xlabel=$F1_{weighted}$,
    tick label style ={/pgf/number format/fixed},
    legend style={at={(0.5,-0.20)},anchor=north},
]
\addplot +[
    hist={
        bins=7,
        data min=0,
        data max=1
    }   ,fill opacity=0.5
] table [y index=0] {wordnet_f1_transfer.csv};
\addplot +[
    hist={
        bins=7,
        data min=0,
        data max=1
    }  ,fill opacity=0.5
] table [y index=0] {doc_f1_transfer.csv};
\legend{WordNet2Vec,Doc2Vec}
\end{axis}
\end{tikzpicture}
\caption{Histogram of $F1_{weighted}$ in~transfer learning for both approaches.}
\label{fig:f1-histogram}
\end{figure}
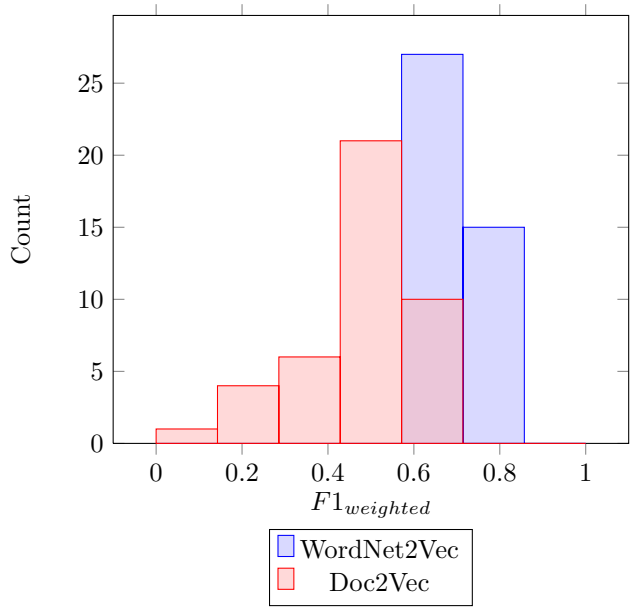

The results achieved in transfer learning setting show the true power and abilities of WordNet2Vec method. We can observe that our method achieves better results than Doc2Vec (Figure~\ref{fig:transfer-doc}, \ref{fig:transfer-wordnet}). It is~important to notice that differences in~results are much bigger in~a~favor of WordNet2Vec in~comparison to results from standard classification (Figure~\ref{fig:transfer-diff}).  The analysis of variance of transfer learning results for both methods shows that WordNet2Vec is~more stable and the results are less various in~comparison to Doc2Vec (Figure~\ref{fig:f1-histogram}). Additionally, we have used statistical tests in~order to check statistical significance of differences in~results (Table~\ref{tab:wilcoxon-transfer}). The superiority of WordNet2Vec over Doc2Vec is~statistically significant. 

\begin{table}[t]
\renewcommand{\arraystretch}{1.3}
\caption{Wilcoxon rank-sum test results for transfer learning.}
\label{tab:wilcoxon-transfer}
\centering
\begin{tabular}{l|c|c|r}
Method 1 & Method 2 & $H_a$ & p-value \\ \hline
Doc2Vec & WordNet2Vec & $F1_{WordNet2Vec}>F1_{Doc2Vec}$ & $6.821*10^{-13}$ \\
\end{tabular}
\end{table}

\section{Conclusions and Future Work}
\label{sec:conclusions_future}

A~novel method---WordNet2Vec---for word vectorization that enables to build more general knowledge representation of texts using WordNets was presented in the paper. It provides a~word representation in~the vector space using its distance to any other word in~the network. In order to present the pair-wise word distance, the method calculates all-pairs shortest paths in~WordNet. 

The usefulness of the WordNet2Vec method was demonstrated in~sentiment analysis problem, i.e. classification with transfer learning setting using Amazon reviews dataset. We compared WordNet2Vec-based classification of sentiment to Doc2Vec approach. Doc2Vec proved to be more accurate in homogeneous setting (learning and testing within the same domain). However, in case of cross domain application (transfer learning), our method outperformed the Doc2Vec results. Hence, we presented its generalization ability in~text classification problems.

In the future work, we want to investigate the different methods of combining word vectors into documents, treat WordNet as multiplex network while calculating shortest paths, reduce feature space of WordNet2Vec Matrix and validate our method on different sources of data such as Twitter or Facebook.


\textbf{Acknowledgments:}  The work was partially supported by The National Science Centre, decision no. DEC-2013/09/B/ST6/02317 and the European Commission under the 7th Framework Programme, Coordination and Support Action, Grant Agreement Number 316097, Engine project. This work was partially supported by the Faculty of Computer Science and Management, Wrocław University of Science and Technology statutory funds.The calculations were carried out in~Wroclaw Centre for Networking and Supercomputing (http://www.wcss.wroc.pl), grant No 177.

\section*{References}

\bibliography{mybibfile}

\end{document}